# ELEMENT: Multi-Modal Retinal Vessel Segmentation Based on a Coupled Region Growing and Machine Learning Approach

Erick O. Rodrigues, Aura Conci, and Panos Liatsis, *Senior Member, IEEE*

***Abstract*—Vascular structures in the retina contain important information for the detection and analysis of ocular diseases, including age-related macular degeneration, diabetic retinopathy and glaucoma. Commonly used modalities in diagnosis of these diseases are fundus photography, scanning laser ophthalmoscope (SLO) and fluorescein angiography (FA). Typically, retinal vessel segmentation is carried out either manually or interactively, which makes it time consuming and prone to human errors. In this research, we propose a new multi-modal framework for vessel segmentation called ELEMENT (vEsseL sEgmentation using Machine lEarning and coNnecTivity). This framework consists of feature extraction and pixel-based classification using region growing and machine learning. The proposed features capture complementary evidence based on grey level and vessel connectivity properties. The latter information is seamlessly propagated through the pixels at the classification phase. ELEMENT reduces inconsistencies and speeds up the segmentation throughput. We analyze and compare the performance of the proposed approach against state-of-the-art vessel segmentation algorithms in three major groups of experiments, for each of the ocular modalities. Our method produced higher overall performance, with an overall accuracy of 97.40%, compared to 25 of the 26 state-of-the-art approaches, including six works based on deep learning, evaluated on the widely known DRIVE fundus image dataset. In the case of the STARE, CHASE-DB, VAMPIRE FA, IOSTAR SLO and RC-SLO datasets, the proposed framework outperformed all of the state-of-the-art methods with accuracies of 98.27%, 97.78%, 98.34%, 98.04% and 98.35%, respectively.**

## I. Introduction

According to the World Health Organisation (WHO), there is approximately 314 million people with visual impairment worldwide, of whom 45 million are blind. Vessel segmentation is a fundamental step in subsequent processing and analysis of retinal images. For instance, segmentation may support registration [1], [2] of retinal images (e.g., in video fluoroscopy), localization of the optic disc and fovea, and aid the analysis of blood flow [3]. In practice, the vasculature of the eye is manually segmented by experts, which is a mundane and laborious task. Moreover, it requires expertise and a considerable degree of attention and time. However, even in the case of clinical experts, retinal vessel annotation is prone to human errors, lacking repeatability and reproducibility. Imaging of retinal structures, specifically the vascular network, is a very complex task for a variety of reasons. In the case of healthy individuals, a multitude of factors, including uneven illumination, the optics of the eye itself interfering with the imaging process, the presence of often multi-layered structures (e.g., macula, fovea, etc) and the very nature of vessels, hinder the development of automated retinal vessel segmentation methods. The task becomes even harder in the presence of retinal pathologies, e.g., exudates and haemorrhages. Given the magnitude of the technical challenge, it is not surprising that there has been a large number of works in the field. In this research, we capitalize on a broad, yet characteristic, set of grey level statistical and connectivity features and the adaptive nature of machine learning to alleviate ambiguities in identifying the retinal vasculature. Indeed, regional image characteristics such as varying contrast and noise may be mitigated by an automated approach. Moreover, vessel narrowings due to the geometric properties of the vasculature or stenoses, and the presence of exudates due to diabetes can be efficiently identified with the aid of machine learning.

This work builds upon principles from image processing and machine learning to propose a novel framework, i.e., ELEMENT, for multi-modal vessel segmentation. The first contribution of this research is the use of connectivity features in conjunction with region growing to identify potential vessel pixels. Next, a complementary set of features for each pixel is proposed, which combines connectivity, and grey level information, i.e., Hessian, Laplacian, Intensity, Anisotropic Diffusion, and Mathematical Morphology-based. Finally, classifiers from the Weka framework [4] are evaluated in terms of their ability to learn a suitable predictive model for segmentation of retinal vasculature. The performance of the proposed approach is evaluated on three types of vessel modalities, namely, Fundus Photography, Scanning Laser Ophthalmoscope and Fluorescein

Erick O. Rodrigues is with the Academic Department of Informatics, Universidade Tecnológica Federal do Paraná (UTFPR), Pato Branco 80230-901, Brazil (e-mail: erickr@id.uff.br).

Aura Conci is with the Department of Computer Science, Universidade Federal Fluminense (UFF), Niteroi 24220-900, Brazil (e-mail: aconci@ic.uff.br).

Panos Liatsis is with the Department of Electrical Engineering and Computer Science, Khalifa University, Abu Dhabi 127788, United Arab Emirates (e-mail: panos.liatsis@ku.ac.ae).

Angiography, and is shown to produce high quality results, due to its adaptive nature. The resulting segmentations are shown to outperform those of a wide range of methods in the state-of-the-art, including deep learning approaches. Moreover, analysis of the results provides supporting evidence for the importance of connectivity features, which are ranked first in all of the experiments.

To the best of our knowledge, this is the first work that approaches the problem of vessel segmentation from this perspective, i.e., incorporating connectivity features and generating a predictive model through the combination of region growing and machine learning.

The remainder of this article is organized as follows. In the next section, we provide an overview of the state-of-the-art, focusing on previous works on vessel enhancement and segmentation in retinal imaging. In Section III, we describe the proposed methodology in detail. Section IV presents the results for the three ocular imaging modalities, as applied to the 6 datasets, provides comparisons with state-of-the-art methods and a broad feature ranking analysis. Finally, Sections V and VI summarize the main aspects of this research, discuss the obtained results and propose avenues for future work.

## II. Literature Review

We will now shift our attention to a brief overview of computer-based methodologies for enhanced visualization of blood vessels and analysis of potential vessel abnormalities. Stergiopulos *et al.*, [5] focus on modeling blood flow within the vessels to automatically locate and quantify the degree of severity of potential stenoses. However, prior to this modeling, vessels are manually segmented. A possible approach for segmenting vessels is the use of filters that enhance vesselness coupled with heuristics for thresholding the filter output.

Although some methods such as the widely popular Frangi filter proposed in [6] enhance visualization of vessels in a broad number of modalities, such as X-ray and Fundus images, most techniques focus on a specific modality and a single group of vessels [7]–[12] in order to achieve better enhancement or segmentation. While the Frangi filter is applicable to various modalities, it is not generally used alone as it does not provide vessel segmentation per se. Application-specific approaches, tightly related to the modality under consideration are used instead.

Sanchez *et al.*, [13] combine Gabor filters with the Boltzmann Univariate Marginal Distribution algorithm to segment vessels in X-ray cardiac angiograms and retinal fundus images. Their methodology consists of optimizing the parameters of the single-scale Gabor filter. The authors use the Area under the Receiver Operating Characteristic Curve (ROC AUC) as the fitness function. Next, they decide whether a pixel belongs to a vessel using interclass variance thresholding.

Odstrcilik *et al.*, [10] proposed a complex method for retinal vessel segmentation in fundus images. At first, the brightness and contrast of the images are corrected. Next, all other channels apart from the green are discarded. Matched filtering is then applied to the preprocessed images using two-dimensional filters. Thereafter, the images are convolved with each of these kernels, which are rotated in various orientations. The resulting parametric images are fused so that the locally maximum response for each pixel is selected. The fused image is thresholded to obtain a binary map of the vessel tree. Finally, the map is further refined using morphological operators [14], removing small artifacts due to presence of noise. The authors compare their methodology to 12 other works. It is possible to infer that their methodology is at least as good as these previously published works.

Region growing is a popular technique used in vessel segmentation [15]–[17]. It is based on grouping pixels or voxels according to a measure of homogeneity. Jiang *et al.*, [17] used an improved region growing method to segment retinal vessels. At first, they enhance the fundus images using the Hessian filter. Next, the Fourier transform coupled with feature detection is used to select seed points. Given the seed points, a set of pre-defined heuristics is used to control region growing. The authors claim that their approach achieves better results than traditional region growing approaches.

Lupascu *et al.*, [18] proposed a method that shares some concepts with the approach proposed in this research. The authors extract feature vectors containing a total of 41 features from the green layer of the fundus images. Their feature vector consists of (1) twenty two region- and boundary-related features in Gaussian scale space, (2) six model-based vessel likelihood features, (3) two features based on the Frangi filter, (4) three features based on Lindeberg ridges, (5) one feature based on Staal ridges, (6) one feature based on the Wavelet transform at multiple scales, (7) eight features based on second-order detectors for elongated structures, and (8) one feature based on intensity-related vesselness likelihood. An Adaboost meta-classifier is trained using previously segmented images. They compare their methodology to 8 works, demonstrating its superiority. Their approach differs from the current framework in that they do not consider the use of region growing in conjunction with the pixel classification task. Another difference lays on the fact that the utilized feature set is substantially different to the one proposed in this work. Our feature set includes more texture-sensitive features, while Lupascu *et al.*, [18] mainly focus on ridge information.

Six works [19]–[24] propose the use of deep learning/convolutional neural networks, which is considered state-of-the-art for a vast number of tasks in image processing. These considered the DRIVE dataset in their experiments, hence we compare the results of the proposed method to theirs. Despite the dominance of deep learning, the results of applying the ELEMENT approach demonstrate higher accuracy rates due to the use of connectivity information.

## III. The ELEMENT Methodology

ELEMENT consists of two main steps: (1) feature extraction, including connectivity features, and (2) region growing, coupled with machine learning-driven pixel classification. The candidate classifiers are trained using feature vectors extracted from images with ground truth, which includes manual per-pixel annotations (i.e., vessel or non-vessel). The first step involves classifier training using ground truth, which produces a predictive/classification model, stored for further usage. The second step consists of locating one or more seed points to commence processing.

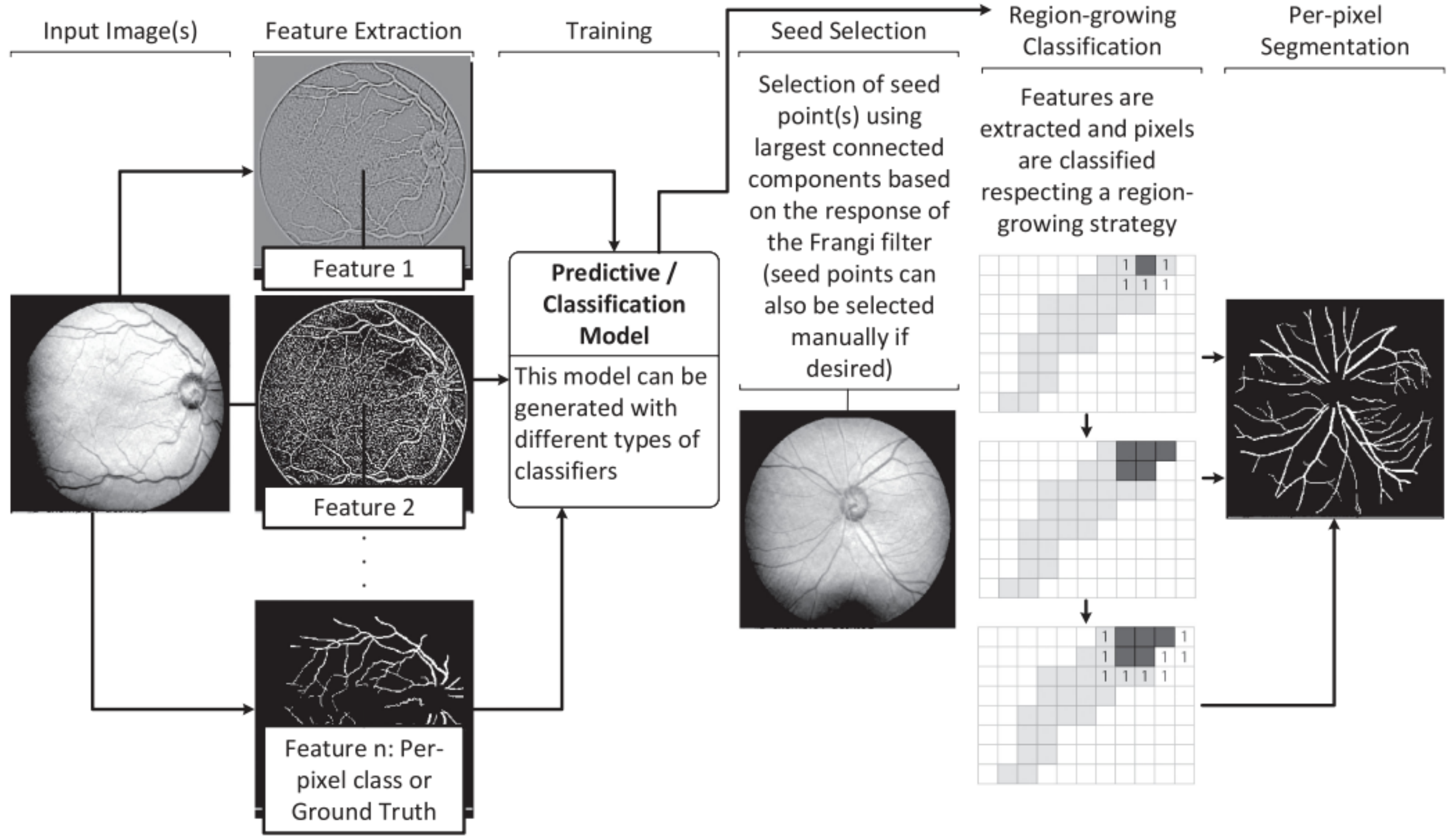

Fig. 1. ELEMENT framework for multi-modal vessel segmentation.

In this work, we select seed points automatically, according to the maximal connected response of the Frangi filter. Specifically, the Frangi filter is applied to the image under consideration and the largest connected component (i.e., the component with the largest number of connected vessel pixels) is determined from Frangi's filter response. The seed points are chosen from this pixel set, prior to segmentation.

Although automated seed selection is employed in our simulation studies, seed points may also be manually selected, i.e., any pixel or group of pixels belonging to the vascular tree would suffice. The use of seed points enhances the odds of correctly classifying neighboring vessel pixels, providing the context for extraction of the connectivity features, and the per-pixel classification process. A schematic of the overall system methodology is illustrated in Fig. 1.

## A. Feature Vector

Vessels adhere to the connectivity property, i.e., there is a higher probability that pixels that are neighbors of vessel pixels belong to the vessel structure. Although conceptually simple, this is a valuable observation, which has not as yet been exploited in vessel segmentation approaches. Most state-of-the-art techniques process the image as a whole, disregarding this important piece of evidence. Specifically, a classification decision on the current pixel does not influence the subsequent classification decisions of its neighbors.

Instead, connectivity is a central principle in region growing. However, region growing does not incorporate the adaptive properties of machine learning methods. There is evidence that region growing alone does not produce sufficiently accurate results [16], [17], [25]. The ELEMENT framework seamlessly incorporates region growing and machine learning, thus combining their respective strengths.

On a similar note, coupling connectivity features with grey-level pattern information provides a rich information vector for classification. The work of Lupascu *et al.*, [18], for instance, performs classification using grey-level pattern information only. In other words, they do not exploit the influence of connectivity information.

Connectivity features provide information that defines whether a pixel is in the vicinity of another vessel pixel, while grey-level features assist in identifying border pixels. In the proposed framework, pixels belonging to vessel structures are updated in real time as the per-pixel classification process advances. Classification occurs in a region growing fashion, which also improves the results in terms of time efficiency and accuracy.

As an example, it is possible to generate simple classification rules based on this complementary feature set. For instance, if a pixel is near a vessel pixel and is not a border pixel, it could be labeled as a candidate vessel pixel. Whereas, if a pixel is in the neighborhood of a vessel pixel and is a border pixel, then it could be labeled as candidate non-vessel pixel. Moreover, when a pixel is not near a vessel pixel, it could be labeled as a candidate non-vessel pixel. Clearly, the rule extraction performed by the machine learning classifiers is far more abstract and complete than this naive analogy. The means to distinguish between border/non-border, and vessel/non-vessel pixels is intrinsic to the properties of each classifier and depends on the learnt rules and paradigms of the model at hand (e.g., decision trees, neural networks, etc).

In what follows we present the feature set used in the multi-modal vessel segmentation framework. First, we discuss connectivity features, followed by a succinct description of grey-level (textural) features, which support detection of border pixels.

*1) Connectivity Features (2 Features):* Two connectivity features are extracted from the pixels to be segmented. The immediate connectivity feature considers the 8-neighbors of the current pixel, and is calculated through a probability function, which relates to the number of neighboring pixels belonging to the vessel. The radial connectivity feature considers a circular area, centered at the current pixel, and is intended to account for regions of disconnected vessel pixels.

The probability of a candidate pixel $I_{i,j}$ belonging to the vessel structure is given by:

$$P(I_{i,j}) = \frac{e^{-M_{i,j}} - 1}{Z} \tag{1}$$

where Z is a normalizing constant to ensure that the probability does not exceed 1, and

$$M_{i,j} = \sum_{m,n \in k} S(M_{i,j}, M_{i+m,j+n}) \tag{2}$$

where k denotes a small neighborhood $M_{i,j}$ around the pixel, typically including the 8-neighbors of the pixel, with

$$S(a, b) = \begin{cases} -1, & \text{if } a == b \\ 0, & \text{otherwise} \end{cases} \tag{3}$$

Next, the immediate connectivity value of the candidate pixel $I_{i,j}$ is defined as:

$$c(I_{i,j}) = \begin{cases} 1, & \text{if } P(I_{i,j} > T_c) \\ 0, & \text{otherwise} \end{cases} \tag{4}$$

where threshold $T_c$ is chosen to have a suitably small value (a value of 0.05 was used in all of our experiments).

For radial connectivity, the neighborhood $k$ in Eq. (2) is adjusted to include the neighbors within a certain radius of the pixel under consideration. In this work, we empirically defined this to be equal to 0.7% of the image diagonal.

Connectivity features are updated in real time. That is, if a recently classified pixel is labeled as vessel, then the vessel array, containing all vessel pixels, is updated in real time and this new pixel information guides the progress of the subsequent classification process.

*2) Grey-Level Features (35 Features):* We now turn our attention to each class of grey-level features. As most of these features are well-known, we provide a concise description of the filter parameters and refer readers to the references within. In summary, the feature vector contains information from: (1) the Hessian matrix [26], (2) the Frangi filter [6], (3) the Difference of Gaussians filter, (4) the Laplacian filter [27], (5) intensity statistics, e.g., mean, (6) anisotropic diffusion [28], (7) morphological opening and closing [1], [29], [30], and (8) image gradients.

*a) Hessian matrix (10 features):* Features based on the Hessian matrix are widely used in vessel segmentation, as mentioned in Section 2. The Frangi filter itself uses information from the Hessian matrix. We extract a total of 10 Hessian-based features. Specifically, the Hessian matrix features are the determinant, the four elements of the matrix, its two eigenvalues, the square of the gamma-normalized square eigenvalue difference (with t = 1), the modulus, and finally, the trace of the Hessian matrix.

*b) Frangi filter (3 features):* The Frangi filter, which is based on the Hessian matrix, can be used to detect edges. It is widely used to emphasize vessels in medical images [6]. In the experiments, three different combinations of parameters for the scale factor, $\sigma$, were considered, generating a total of 3 features: $\{\sigma_{start} = 1, \sigma_{end} = 1\}, \{\sigma_{start} = 1, \sigma_{end} = 2\}$, and $\{\sigma_{start} = 2, \sigma_{start} = 3\}$. The parameter $\beta_1$ was set to 2, $\beta_2$ was set to 1 and the $\sigma$-step was set to 1 for all three features. These parameters can be easily adjusted to accommodate high resolution images, if necessary.

*c) Laplacian or sharpen filter (3 features):* The Laplacian is an isotropic measure of the second-order spatial derivative. It is frequently used in the literature as an edge detector as it highlights abrupt intensity changes in images. The Laplacian can be computed using an inverted Gaussian kernel. Two Laplacian filters were applied using kernel sizes of 3x3 and 20x20, respectively, producing a total of 2 features. In addition, a simple non-Gaussian kernel, given by the following matrix: $\{\{-1, -1, -1\}, \{-1, 9, -1\}, \{-1, -1, --\}\}$ was also used, generating the third feature.

*d) Mean, max, min and median values (5 features):* The arithmetic mean, geometric mean, max, min and median values were computed in a square window of $7 \times 7$ pixels, centered on the pixel, producing a total of 5 features, respectively.

*e) Anisotropic diffusion (4 features):* Anisotropic diffusion removes noise from images, while preserving borders, thus being different to the use of traditional blurring filters [31]. Four anisotropic diffusion features were calculated using the following parameters: $\{(i = 10, \kappa = 3, \lambda = 0.5), (i = 20, \kappa = 4, \lambda = 0.3), (i = 40, \kappa = 6, \lambda = 0.8), (i = 35, \kappa = 3, \lambda = 2)\}$.

*f) Morphological opening and closing (6 features):* Opening and closing are operations based on dilation and erosion from mathematical morphology. Opening is defined as the application of dilation followed by erosion, while closing is its dual operation and is defined as the application of erosion followed by dilation [14]. Two types of structuring elements were used in our experiments: $B_1 = \{(0, 255, 0), (255, 255, 255), (0, 255, 0)\}$ and $B_2$, which is a 10x10 mask containing a central symmetrical discrete circle filled with the value of 255. A total of 6 features were extracted. The parameters of these features are as follows: $\{(i = 1, j = 1, B = B_1), (i = 3, j = 1, B = B_1), (i = 1, j = 3, B = B_1), (i = 1, j = 1, B = B_2), (i = 3, j = 1, B = B_2), (i = 1, j = 3, B = B_2)\}$, where $i$ represents the number of times the erosion operation is applied. Similarly, $j$ represents the number of times the dilation operation is applied.

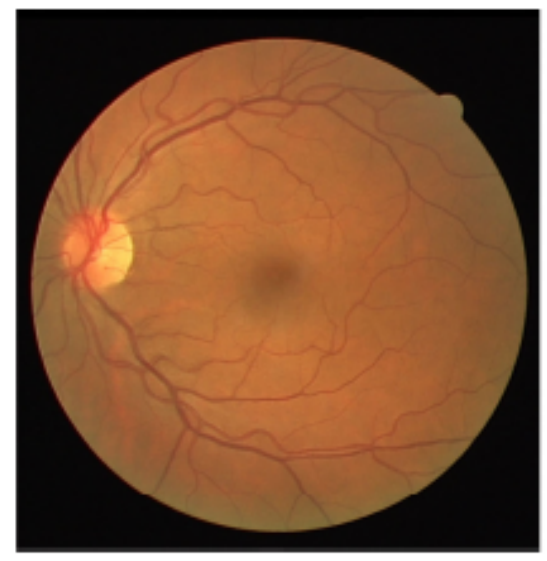

(a) Retinal image with vessels depicted in red.

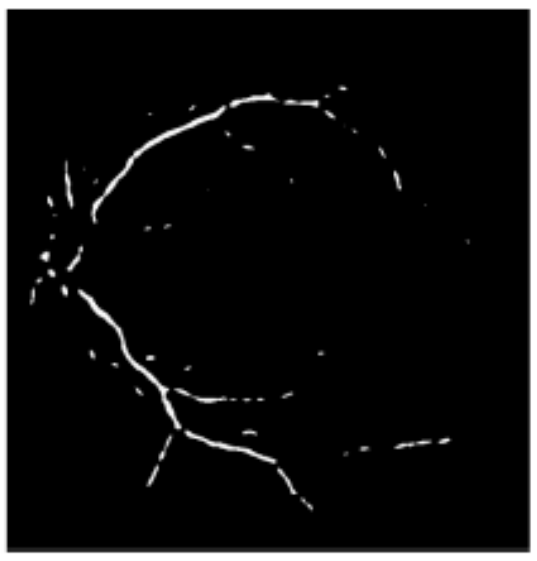

(b) Candidate seed points of (a) - noiseless response of the Frangi filter. The largest connected components are better suited for seed selection.

Fig. 2. Noiseless response of the Frangi filter (potential seed points).

*g) Gradient information (4 Features):* Two gradient features were calculated using two means of gradient information estimation, i.e., a linear and 3 different Gaussian kernels.

## B. Classification and Segmentation Approach

This stage consists of extracting information from manually labeled pixels in order to generate a predictive model. In our modeling process, classification is formulated as a binary problem where the label of each pixel is either vessel or non-vessel.

Initially, a seed point belonging to the vascular tree must be selected. Given image (a) in Fig. 2 and its respective ground truth (b), any pixel that belongs to the ground truth (i.e., white pixels) is a candidate seed point.

As previously mentioned, in this step, we apply the Frangi filter to the input image and select the largest connected component, based on the response of the Frangi filter, as the pixel group for initial seed point selection. The largest connected connected component from Frangi's response is typically the longest and/or thickest vessel structure in the image. The Frangi filter is very sensitive to this information, although it lacks precision in the case of narrow vessels and in the presence of noise, and indeed these cases are better suited for our segmentation approach.

One fundamental design characteristic of the ELEMENT framework is that it exploits the strategy of region growing. This property is inherited through the use of connectivity features, where the classification of the current pixel is partially dependent on the presence or absence of vessel pixels in its vicinity.

Let $V$ be the set of the positions of pixels, classified as vessel pixels and seed points. Set $V$ is updated in real time during the pixel classification process. Candidate pixels for classification are those with minimal distance in respect to pixels in $V$.

In Eq. (5), $V_0$ refers to set $V$ at iteration 0. $V_0$ contains the initially selected seed points. $S$ is the set of all seed points. The classification process is terminated at iteration $i$ if the elements belonging to $V_i$ have been already classified in previous iterations (i.e., all vessel pixels are already classified).

$$V_0 = \{a : a \in S\}$$

$$\vdots$$

$$V_i = \{b : d(b, c) = t, c \in V_{i-1}\} \cup V_{i-1} \tag{5}$$

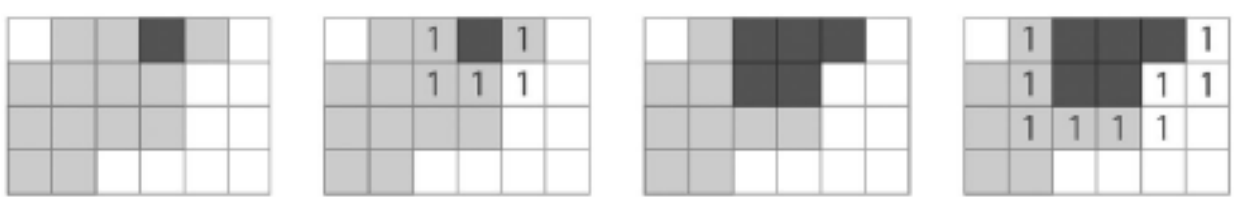

(a) Seed point ($V_0$) in dark grey at iteration 0. (b) Candidate pixels to be classified at distance 1 from the seed point. (c) Pixels marked as vessel ($V_1$) in dark grey at iteration 1. (d) Candidate pixels to be classified at distance 1 with regards to $V_1$.

Fig. 3. Pixel classification process.

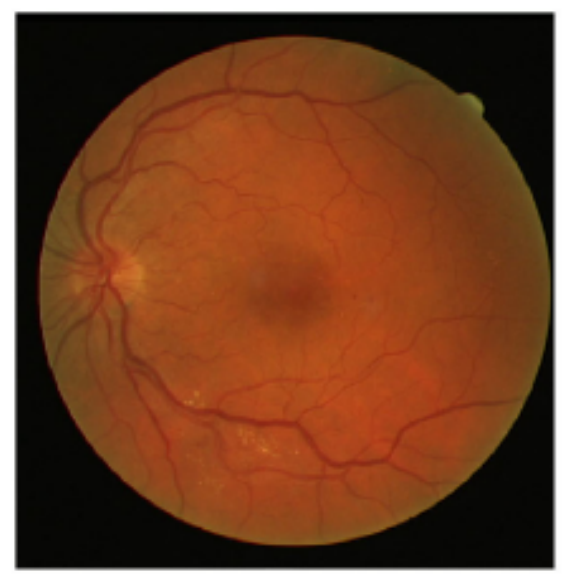

(a) Image 1 - test dataset.

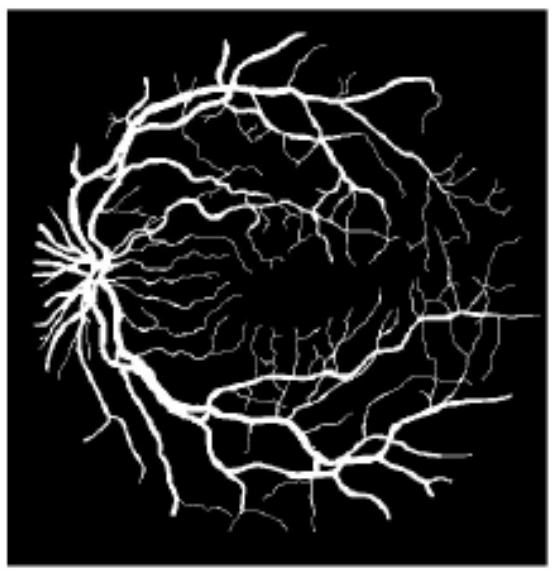

(b) Ground truth of (a).

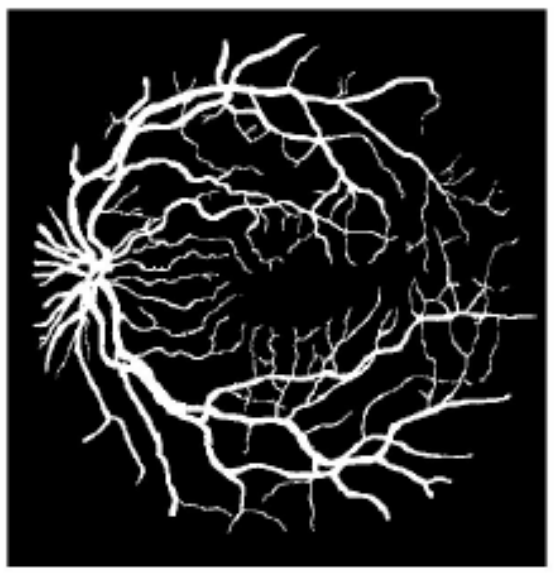

(c) Segmentation of (a) produced by ELEMENT.

Fig. 4. ELEMENT segmentation results on images of the DRIVE dataset exhibiting retinal abnormalities.

Function $d()$ represents the supremum distance metric and $t$ is set to 1. The set of candidate points to be classified at iteration $i$ is given by $V_i \setminus V_{i-1}$. That is, the points that are at distance $t$ in regards to the points in $V_{i-1}$. Fig. 3 illustrates the classification process given a single seed point over two iterations.

In Fig. 3, light grey pixels indicate the location of actual vessel pixels and dark grey pixels indicate seed points and also points in $V_i$. The classification process is applied until the entire image is processed or all pixels in $V_i$ have been previously classified.

Fig. 3(a) illustrates the initial seed point in dark grey ($V_0$). Next, the pixels marked with 1's in 3(b), which are at distance of 1 pixel from $V_0$, are considered for classification. The feature vectors of these candidate pixels are calculated based on the information of the previous classification step and the grey-level information of the pixels. Finally, a classification model, generated using a machine learning algorithm, decides whether the pixel is classified as a vessel pixel. This process is repeated until the image is fully processed.

## C. Classification Algorithms

This section provides a brief overview of the workings of the classification process, instead of focusing on individual classifiers. Further information including fundamental concepts and implementation details can be found in the Weka framework [4], [32] and the references therein.

TABLE I
EXPERIMENTAL DATASETS AND ASSOCIATED DETAILS

| | | |
|---|---|---|
| DRIVE | DATASET SIZE | 40 (20 train, 20 test) |
| | IMAGE RESOLUTION | 565x584 |
| | TRAINED WITH | image 21 (train set) |
| | TESTED WITH | 20 images (test set) |
| | MODALITY | fundus images |
| STARE | DATASET SIZE | 20 |
| | IMAGE RESOLUTION | 700x605 |
| | TRAINED WITH | images im0001 and im0236 |
| | TESTED WITH | 18 images (exc. im0001 and im0236) |
| | MODALITY | fundus images |
| CHASE-DB | DATASET SIZE | 28 |
| | IMAGE RESOLUTION | 999x960 |
| | TRAINED WITH | images 05L, 08L and 11L |
| | TESTED WITH | 25 images (exc. 05L, 08L and 11L) |
| | MODALITY | fundus images |
| VAMPIRE | DATASET SIZE | 8 |
| | IMAGE RESOLUTION | 3900x3072 |
| | TRAINED WITH | image amd1 |
| | TESTED WITH | 7 images (exc. amd1) |
| | MODALITY | fluorescein angiography |
| IOSTAR | DATASET SIZE | 24 |
| | IMAGE RESOLUTION | 565x555 |
| | TRAINED WITH | image 02 |
| | TESTED WITH | 23 images (exc. 02) |
| | MODALITY | scanning laser ophthalmoscope |
| RC-SLO | DATASET SIZE | 40 |
| | IMAGE RESOLUTION | 320x280 |
| | TRAINED WITH | images STAR05Patch3, STAR06Patch1 and STAR17Patch4 |
| | TESTED WITH | 37 images (exc. STAR05Patch3, STAR06Patch1 and STAR17Patch4) |
| | MODALITY | scanning laser ophthalmoscope |

The framework proposed in this research decides whether a pixel belongs to the background or the vasculature, i.e., vessel segmentation is posed as a binary classification problem. The classifier is trained using pixels, which were manually segmented (i.e., ground truth). Each of the classifiers was trained according to the details of Table I. During this phase, grey-level information from the image is extracted and combined with the manual annotations (i.e., belonging or not to a vessel structure). The model, generated by the classifier, learns to associate specific feature patterns to labels during training and uses this information to produce a label, when it is unknown, i.e., in the testing phase.

The following classifiers from the Weka framework have been evaluated: (i) Random Forest, (ii) Radial Basis Function (RBF) networks, (iii) stochastic variant of Pegasos (SPegasos), (iv) fast decision tree with reduced error pruning (REPTree), (v) Naive Bayes, (vi) Bayes Net, (vii) grafted decision tree using the J48 graph method (J48 Graft), (viii) Decision Table, and (ix) Hoeffding Tree.

In the following section, we perform the extraction of the previously described features and carry out experiments with the Weka framework classifiers. Finally, we report on the performance rates obtained with the best classifier for each of the ocular modality datasets.

## IV. EXPERIMENTAL RESULTS

In this section, we present six sets of experiments comprising a total of three modalities. The first modality is retinal fundus images and is represented by the widely known DRIVE, STARE and CHASE-DB datasets. The second modality is fluorescein angiography (FA) retinal images, where the VAMPIRE dataset is used. Finally, the third modality is scanning laser ophthalmoscope (SLO), where the IOSTAR and RC-SLO datasets are used. Table I shows the details of each dataset, including the training and test datasets.

### A. Retinal Fundus (DRIVE Dataset)

The DRIVE dataset resulted from a diabetic retinopathy screening program in the Netherlands [33]. The screening population consisted of 400 diabetic subjects between 25-90 years of age. Forty photographs were randomly selected, where 33 do not show any signs of diabetic retinopathy and 7 show signs of mild early diabetic retinopathy. Fig. 4 shows two images of the DRIVE dataset (a) and (d), their ground truth, (b) and (e), respectively, and the segmentation results produced by the proposed framework, (c) and (f), respectively. The experiments were performed using the green channel only as it is richer in terms of information for this particular modality (i.e., the red and blue layers were excluded).

In this case, classifiers were trained using a single image from the DRIVE dataset (specifically, the first image in the training folder). A single image in the DRIVE dataset has approximately 180,000 pixels. One image was deemed sufficient for training in this dataset, as there is not a lot of variation among images. However, increasing the size of the training set could lead to subtle improvements on the segmentation results at the expense of training times. ELEMENT was tested on the DRIVE test folder (20 images) as it is the standard for this dataset and the rates provided in Table II refer to the test folder. It is important to note that Fig. 4(a) is obtained from a patient who exhibits diabetic retinopathy.

Table II compares the results obtained with each classifier. Random Forest achieved the highest Receiver Operating Curve (ROC) - Area Under the Curve (AUC) and accuracy. The Radial Basis Function (RBF) classifier, however, obtained a notably high accuracy, while requiring lower training and testing times in comparison to Random Forest. Three manually selected combinations of parameters were evaluated for each classifier and the best combination was selected.

The best result obtained in this set of experiments (i.e., Random Forest classifier) was compared to several state-of-the-art vessel segmentation approaches, which use the DRIVE dataset as a means of performance evaluation. This comparison is shown in Table III. The work of Sheet *et al.*, [34] is the only one that achieved a higher accuracy (97.66%) in comparison to ELEMENT (97.4%). However, their method results to a grey-level vessel probability map, which further requires the selection of a

TABLE II
COMPARISON OF CLASSIFICATION ALGORITHMS FOR THE DRIVE DATASET

| Classifier | TP (%) | TN (%) | Accuracy (%) | AUC | Train Time (s) | Test Time (s) |
|---|---|---|---|---|---|---|
| Random Forest | 89.83 | 98.16 | **97.4** | **0.9936** | 1613.96 | 89.94 |
| RBF Classifier | 91.78 | 97.83 | 97.3 | 0.9915 | 120.1 | 4.93 |
| SPegasos | 68.33 | 98.57 | 95.93 | 0.8346 | 46.37 | 4.31 |
| REPTree | 90.68 | 97.74 | 97.1 | 0.9849 | 41.5 | 4.02 |
| Naive Bayes | 85.81 | 87.6 | 87.1 | 0.9361 | 28.35 | 13.12 |
| Bayes Net | **97.78** | 81.75 | 83.15 | 0.9733 | 67.81 | 8.1 |
| J48 Graft | 87.86 | 97.97 | 97.08 | 0.9297 | 120.25 | **3.74** |
| Decision Table | 92.39 | 97.78 | 97.3 | 0.9916 | 1850.08 | 31.6 |
| Hoeffding Tree | 69.31 | **98.96** | 96.36 | 0.9089 | **26.41** | 26.63 |

* TP/TN refer to true positive/negative rate, respectively.
* Accuracy is given by: (true positives + true negatives) / total population.
* AUC refers to the area under the Receiver Operating Characteristic (ROC) curve.

TABLE III
COMPARISON OF RESULTS WITH THE STATE-OF-THE-ART (DRIVE DATASET)

| Work | TP | TN | ACC | AUC | F1 score | MCC |
|---|---|---|---|---|---|---|
| Staal et al., [33] | - | - | 94.41 | 0.9516 | - | - |
| Jiang et al., [33], [35] | - | - | 89.11 | 0.9009 | - | - |
| Sheet et al., [34] | - | - | **97.66** | - | - | - |
| Lupascu et al., [18] | - | - | 95.97 | 0.9561 | - | - |
| Al-rawi et al., [36] | - | - | 94.2 | 0.9582 | - | - |
| Soares et al., [37] | - | - | 94.66 | 0.9614 | - | - |
| Marin et al., [38] | 70.67 | 98.01 | 94.52 | 0.9588 | - | - |
| Lam et al., [39] | - | - | 94.72 | 0.9614 | - | - |
| Zhang et al., [40] | 71.2 | - | 93.82 | - | - | - |
| Delibasis et al., [41] | 67.31 | 97.58 | 93.77 | - | - | - |
| Odstrcilik et al., [10] | 70.6 | 96.93 | 93.4 | 0.9519 | - | - |
| Al-Diri et al., [42] | 72.82 | 95.51 | - | - | - | - |
| Perez et al., [25] | 64.4 | - | 92.5 | - | - | - |
| Mendonca et al., [43] | 73.15 | - | 94.52 | - | - | - |
| Ricci et al., [44] | - | - | 95.95 | 0.9633 | - | - |
| Maji et al., [24] | - | - | 93.27 | - | - | - |
| Azzopardi et al., [45] | 76.55 | 97.04 | 94.42 | 0.9614 | - | 0.7188 |
| Zhang et al., [46] | 77.43 | 97.25 | 94.76 | 0.9636 | - | - |
| Li et al., [47] | 75.69 | 98.16 | 95.27 | 0.9738 | - | - |
| Fu et al., [23] | 72.94 | - | 94.70 | - | - | - |
| Hossain et al., [48] | 78.63 | 97.11 | - | - | - | - |
| Liskowski et al., [22] | 87.03 | **99.29** | 95.15 | 0.9515 | - | - |
| Z. Fan et al., [49] | 71.90 | 98.50 | 96.10 | - | - | - |
| Yan et al., [19] | 76.53 | 98.18 | 95.42 | 0.9752 | - | - |
| J. Mo et al., [20] | 77.79 | 97.80 | 95.21 | - | - | - |
| Welikala et al., [21] | - | - | 91.99 | - | - | - |
| Jin et al., [50] | 78.94 | 98.70 | 96.97 | 0.9855 | 0.8203 | - |
| Srinidhi et al., [51] | 86.44 | 96.67 | 95.89 | 0.9701 | 0.7607 | 0.7421 |
| Orlando et al., [52] | 78.97 | 96.84 | - | - | 0.7857 | 0.7556 |
| Samant et al., [53] | 81.45 | 98.66 | 96.96 | 0.8137 | - | 0.7870 |
| Y. Jiang et al., [54] | 78.39 | 98.90 | 97.09 | 0.9864 | 0.8246 | - |
| Soomro et al., [55] | 87.00 | 98.5 | 95.6 | - | - | - |
| Adeyinka et al., [56] | 76.03 | - | 95.23 | - | - | - |
| Biswal et al., [57] | 71.00 | 97.00 | 95.0 | - | 0.75 | 0.76 |
| Chakraborti et al., [58] | 72.05 | 95.79 | 93.70 | - | - | - |
| Biswal et al., [59] | 78.00 | 98.00 | 97.0 | - | 0.83 | 0.80 |
| Shin et al., [60] | 92.55 | 93.82 | 92.71 | 0.9802 | - | - |
| ELEMENT | **89.83** | 98.16 | *97.4* | **0.9936** | **0.8579** | **0.8461** |

* TP refers to true positive rate and TN refers to true negative rate.
* Accuracy (ACC) is given by: (true positives + true negatives) / total population.
* AUC refers to the area under the Receiver Operating Characteristic (ROC) curve.
* F1 refers to the F1-score.
* MCC refers to the Matthews Correlation Coefficient.

TABLE IV
COMPARISON OF RESULTS WITH THE STATE-OF-THE-ART (STARE DATASET)

| Work | TP | TN | ACC | AUC | F1 | MCC |
|---|---|---|---|---|---|---|
| Roychowdhury et al. [63] | - | - | 95.35 | 0.963 | - | - |
| Hoover et al., [61] | 80.00 | 90.00 | - | - | - | - |
| Mendonça et al. [43], | 67.64 | - | 94.79 | - | - | - |
| Soares et al., [37] | - | - | 94.80 | 0.967 | - | - |
| Marin et al., [38] | - | - | 95.26 | 0.976 | - | - |
| Lam et al., [64] | - | - | 94.74 | 0.939 | - | - |
| Annunziata et al., [65] | 71.28 | 98.36 | 95.62 | 0.965 | - | - |
| Azzopardi et al., [45] | 77.16 | 97.01 | 94.97 | 0.956 | - | 0.733 |
| Ricci et al., [44] | - | - | 96.46 | 0.968 | - | - |
| Zhao et al., [66] | 78.00 | 97.80 | 95.60 | 0.874 | - | - |
| Fraz et al., [67] | 89.51 | 93.84 | 93.48 | - | - | - |
| Yongping et al., [68] | 93.73 | - | 90.87 | - | - | - |
| Mo et al., [20] | 81.47 | 98.44 | 96.74 | 0.988 | - | - |
| Lupascu et al., [18] | - | - | 95.97 | 0.956 | - | - |
| Liskowski et al., [22] | 89.66 | 84.50 | 97.40 | **0.994** | - | - |
| Imani et al., [69] | 75.02 | 97.45 | 95.90 | - | - | - |
| Jin et al., [50] | 84.19 | 95.63 | 94.45 | 0.969 | - | - |
| Srinidhi et al., [51] | 83.25 | 97.46 | 95.02 | 0.967 | 0.769 | 0.739 |
| Orlando et al., [52] | 76.80 | 97.38 | - | - | 0.764 | 0.741 |
| Odstrcilik et al., [10] | 78.47 | 95.12 | 93.41 | 0.934 | - | - |
| Ricci et al., [44] | - | - | 96.80 | 0.964 | - | - |
| Li et al., [47] | 70.27 | 98.28 | 95.45 | 0.954 | - | - |
| Z. Fan et al., [49] | 70.00 | 97.9 | 95.90 | - | - | - |
| Samant et al., [53] | 68.69 | 98.16 | 95.94 | 0.9253 | - | 0.7682 |
| Y. Jiang et al., [54] | 82.49 | 99.04 | 97.81 | 0.9927 | 0.8482 | - |
| Asad et al., [70] | 74.83 | 95.44 | 93.39 | - | - | - |
| Soomro et al., [55] | 84.80 | 98.60 | 96.80 | - | - | - |
| Adeyinka et al., [56] | 74.12 | - | 95.85 | - | - | - |
| Biswal et al., [57] | 70.0 | 97.0 | 95.0 | - | 0.76 | 0.74 |
| Chakraborti et al., [58] | 67.86 | 95.86 | 93.79 | - | - | - |
| Biswal et al., [59] | 80.0 | 96.0 | 96.0 | - | 0.82 | 0.78 |
| Shin et al., [60] | 93.52 | 95.98 | 93.78 | 0.9877 | - | - |
| ELEMENT | **94.26** | **98.62** | **98.27** | **0.9946** | **0.8910** | **0.8834** |

* TP refers to true positive rate and TN refers to true negative rate.
* Accuracy (ACC) is given by: (true positives + true negatives) / total population.
* AUC refers to the area under the Receiver Operating Characteristic (ROC) curve.
* F1 refers to the F1-score.
* MCC refers to the Matthews Correlation Coefficient.

threshold value. That is, their approach does not directly result in vessel segmentation. In contrast, the ELEMENT framework produces a binary image highlighting pixels that are classified as belonging to the vasculature. Furthermore, their approach is only tested on retinal fundus images, in contrast to the proposed multi-modal framework.

Overall, the proposed methodology achieved the highest true positive rate, true negative rate and AUC. AUC indicates that ELEMENT separates the two classes (vessel and non-vessel) significantly better than the existing approaches over all possible threshold values. The true positive rate, on the other hand, is one of the most important metrics as it is related to the detection of vessel pixels. An entirely black image would have the highest true negative rate but 0% true positive rate.

As a remark, for the remaining analyses shown in this work, the Random Forest classifier will be used due to its superior performance and generalization capabilities.

### B. Retinal Fundus (STARE Dataset)

The STARE (STructured Analysis of the REtina) project was conceived and initiated in 1975 by Michael Goldbaum, M.D.,

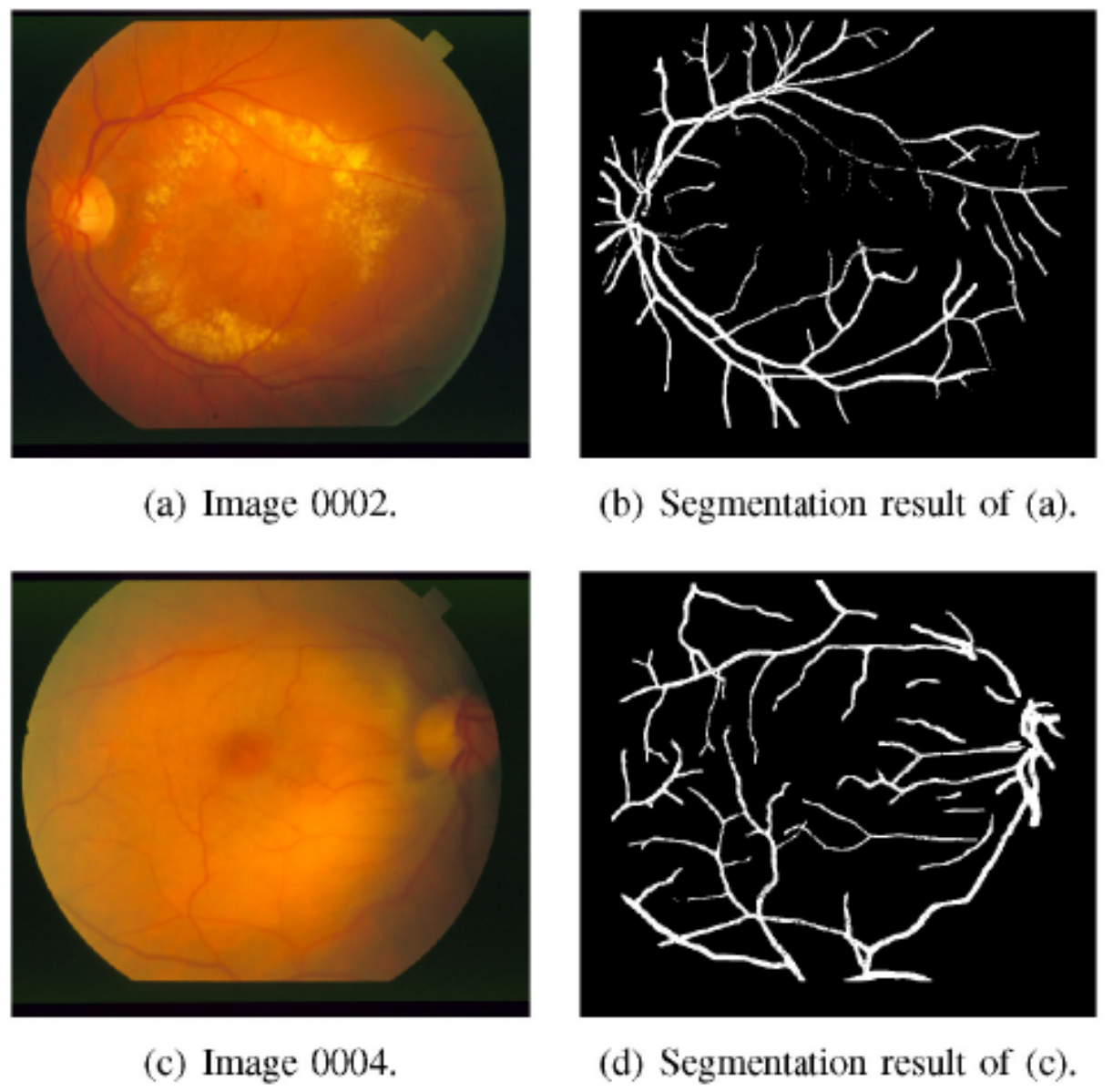

(a) Image 0002. (b) Segmentation result of (a).

(c) Image 0004. (d) Segmentation result of (c).

Fig. 5. ELEMENT segmentation results on images of the STARE dataset exhibiting retinal abnormalities.

at the University of California, San Diego. It was funded by the U.S. National Institutes of Health. Images and clinical data were provided by the Shiley Eye Center at the University of California, San Diego, and the Veterans Administration Medical Center in San Diego [61], [62].

Although the STARE dataset contains a large number of images, only 20 of these images contain vessel annotations [61]. In this case, we used images 0001 and 0236 for training and evaluated the performance of our segmentation on the remaining 18 images. Selecting more than one image for training improves the generalization capabilities of the predictive model, but also reduces accuracy when considering very similar images (i.e., overfitting). In contrast to the DRIVE dataset, STARE contains several images demonstrating retinal abnormalities, thus exhibiting more variation. Therefore, two images were used in training, so as to mitigate for general variations in image information.

Table IV shows how ELEMENT compares to the state-of-the-art in the STARE dataset. The proposed framework was able to outperform all state-of-the-art works.

Fig. 5 shows the results of the proposed segmentation framework on images with retinal pathologies. Figure 5-(a) demonstrates choroidal neovascularization and arteriosclerotic retinopathy, while (c) is indicative of cilio-retinal artery occlusion or central retinal artery occlusion.

## C. Retinal Fundus (CHASE-DB Dataset)

The CHASE-DB dataset was developed during the Child Heart Health Study in England (CHASE), a cardiovascular health survey in 200 primary schools in London, Birmingham, and Leicester. It captured information from 19 pupils from 10 primary schools, who were measured both in the morning and the afternoon on the same day between September 2007 and March 2008 by the same observer.

TABLE V
COMPARISON OF RESULTS WITH THE STATE-OF-THE-ART (CHASE-DB DATASET)

| Work | TP | TN | ACC | AUC | F1 | MCC |
|---|---|---|---|---|---|---|
| Liskowski et al., [22] | 78.16 | 98.36 | 96.28 | 0.9823 | - | - |
| Srinidhi et al., [51] | 82.97 | 96.63 | 94.74 | 0.9591 | 0.7189 | 0.6927 |
| Orlando et al., [52] | 72.77 | 97.12 | - | - | 0.7332 | 0.7046 |
| Samant et al., [53] | 70.27 | 98.28 | 95.45 | 0.9102 | - | 0.7563 |
| Y. Jiang et al., [54] | 78.39 | 98.94 | 97.21 | 0.9866 | 0.8062 | - |
| Soomro et al., [55] | **88.6** | 98.2 | 97.6 | - | - | - |
| Adeyinka et al., [56] | 71.3 | - | 94.89 | - | - | - |
| Biswal et al., [57] | 76.0 | 97.0 | - | - | 0.75 | 0.73 |
| Chakraborti et al., [58] | 52.86 | 95.94 | 92.98 | - | - | - |
| Biswal et al., [59] | 78.0 | 97.0 | 97.0 | - | 0.76 | 0.73 |
| Shin et al., [60] | 93.64 | 94.63 | 93.73 | 0.9830 | - | - |
| ELEMENT | 87.82 | **98.52** | **97.78** | **0.9923** | **0.8448** | **0.8343** |

* TP refers to true positive rate and TN refers to true negative rate.
* Accuracy (ACC) is given by: (true positives + true negatives) / total population. * AUC refers to the area under the Receiver Operating Characteristic (ROC) curve. * F1 refers to the F1-score. * MCC refers to the Matthews Correlation Coefficient.

Table V shows the results obtained with the CHASE-DB dataset. The ground truth generated by the first observer was used. This is yet another case where ELEMENT by and large outperformed the state-of-the-art. In this case, Random Forest was trained using images 05L, 08L and 11L, and its performance tested on the remaining images of the dataset.

## D. Retinal Fluorescein Angiograms (VAMPIRE Dataset)

In order to complement the previous analyses, this subsection addresses a different retinal imaging modality. The Fluorescein Angiography (FA) retinal dataset was used, provided by Rovira *et al.*, [71], which is also known as the VAMPIRE dataset. This is composed of 8 high resolution greyscale images of $3600 \times 3072$ pixels. FA is used to examine blood circulation in the retina using the dye tracing method. As fluorescein dye passes through the retinal vasculature, it reveals abnormalities, such as leakage due to the breakdown of the blood-retina barrier, areas of non-perfusion, occluded vessels and the presence of new and anomalous vasculature, such as microaneurysms, neovascularisation and arteriovenous shunts. Various parts of the vasculature become visible at different times during the course of the FA procedure.

Rovira *et al.*, [71] state that most methods in retinal vessel segmentation involve fundus images, emphasizing that no quantitative evaluation of FA segmentation has been reported. Indeed, their work is the first to provide such quantitative evaluation.

In this set of experiments, image amd1 was used for training purposes, and segmentations were produced in the testing phase of the system for the remaining seven images of the dataset (i.e., amd2 to ger4). In all cases, the Random Forest classifier was used. The complete per-image statistics can be found in [75].

Table VI reports the average accuracy, true positive and negative rates obtained in the VAMPIRE dataset [71]. ELEMENT

TABLE VI
COMPARISON OF RESULTS WITH THE STATE-OF-THE-ART (VAMPIRE DATASET)

| Work | TP | TN | ACC | AUC | F1 | MCC |
|---|---|---|---|---|---|---|
| Perez-Rovira et al, [71] | 58.38 | **99.05** | 97.63 | - | - | - |
| Zhao, Rada et al., [66] | 72.9 | 98.5 | 97.7 | - | - | - |
| Ding et al, [72] | 66.4 | 98.9 | 97.9 | - | - | - |
| Zhao, Liu et al., [73] | 72.1 | 98 | 97.6 | - | - | - |
| Ding et al., [74] | - | - | - | 0.995 | - | - |
| Biswal et al., [59], | 79.0 | 97.0 | 96.0 | 0.88 | - | - |
| ELEMENT | **97.22** | 98.36 | **98.34** | **0.9956** | **0.8935** | **0.8886** |

* TP refers to true positive rate and TN refers to true negative rate. * Accuracy (ACC) is given by: (true positives + true negatives) / total population. * AUC refers to the area under the Receiver Operating Characteristic (ROC) curve. * F1 refers to the F1-score. * MCC refers to the Matthews Correlation Coefficient.

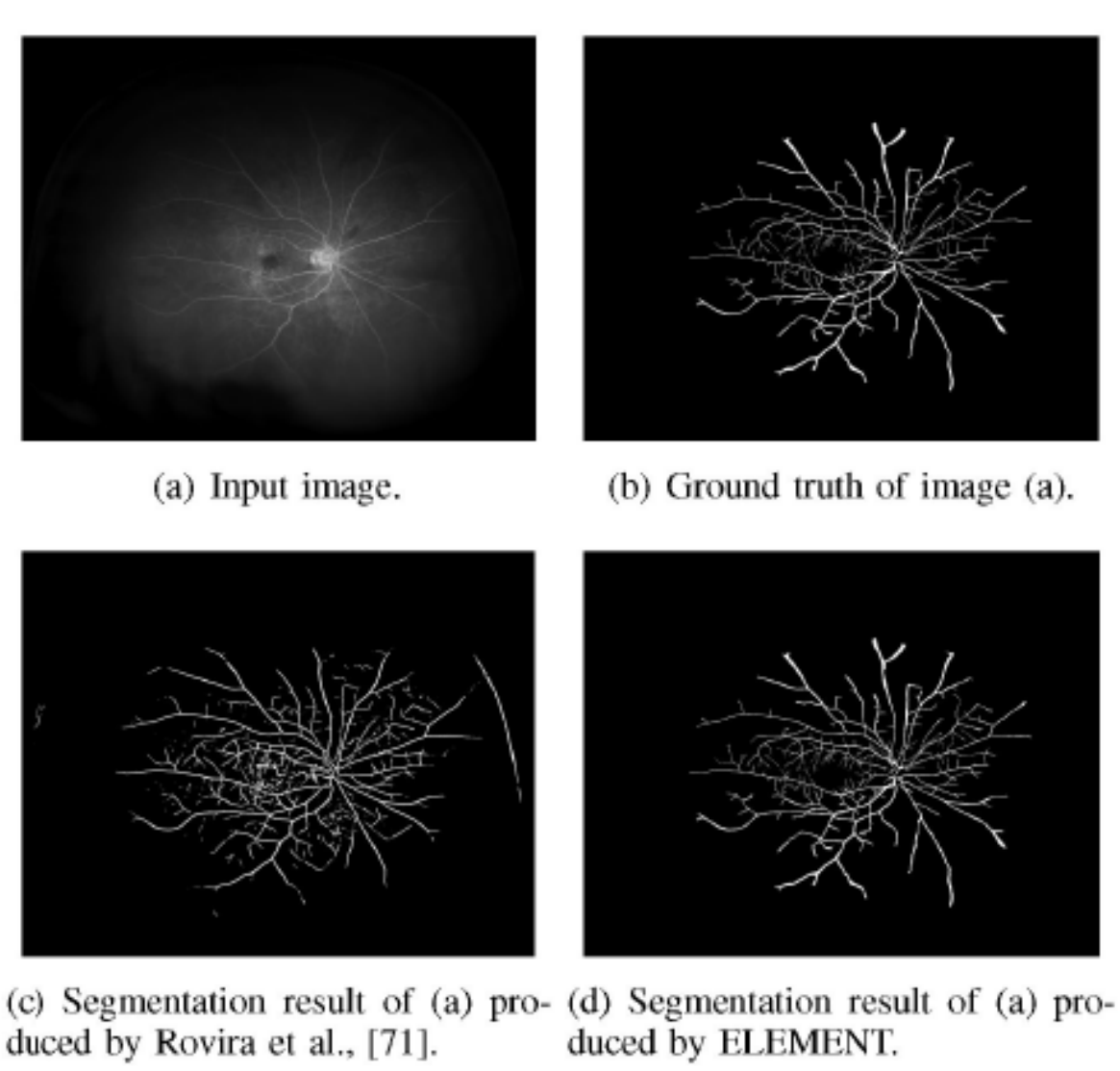

(a) Input image. (b) Ground truth of image (a).

(c) Segmentation result of (a) produced by Rovira et al., [71]. (d) Segmentation result of (a) produced by ELEMENT.

Fig. 6. Retinal fluorescein angiograms, (a) and (e), their ground truth, (b) and (c), respectively, segmented by Rovira *et al.*, [71], (c) and (g), respectively, and the ELEMENT framework, (d) and (h), respectively.

outperforms the state-of-the-art in two of the three statistics (i.e., TP, TN and accuracy). The TP rate is substantially higher, which indicates robust segmentation of vessel pixels. Fig. 6 shows segmentation results produced by Rovira et al. [71], the proposed approach and the associated ground truth.

The recent work of Ding *et al.*, [76] reports an average ROC-AUC of 0.9758 over the 8 images and later, in [74], an average ROC-AUC of 0.995. The proposed method achieved an average ROC-AUC of 0.9956. However, it is not straightforward to measure the quality of an algorithm by just evaluating its ROC-AUC value.

## *E. SLO Retinal Fundus (IOSTAR Dataset)*

The images in the IOSTAR dataset [77] were taken with an EasyScan camera based on the scanning laser ophthalmoscope (SLO) technology (using green and infrared lasers). These images have a resolution of 1024x1024 with a Field of View (FOV)

TABLE VII
COMPARISON OF RESULTS WITH THE STATE-OF-THE-ART (IOSTAR DATASET)

| Work | TP | TN | ACC | AUC | F1 | MCC |
|---|---|---|---|---|---|---|
| Sureshjani et al., [77] | 78.63 | 98.05 | 95.07 | 0.9615 | - | - |
| Soares et al., [37] | 76.76 | 97.2 | 94.61 | 0.9603 | - | - |
| Srinidhi et al., [78] | **88.00** | 84.00 | 89.00 | 0.9400 | 0.7354 | 0.7057 |
| Zhao et al., [79] | 79.15 | 97.92 | - | - | - | - |
| ELEMENT | 86.49 | **98.96** | **98.04** | **0.9959** | **0.8635** | **0.8561** |

* TP refers to true positive rate and TN refers to true negative rate. * Accuracy (ACC) is given by: (true positives + true negatives) / total population. * AUC refers to the area under the Receiver Operating Characteristic (ROC) curve. * F1 refers to the F1-score. * MCC refers to the Matthews Correlation Coefficient.

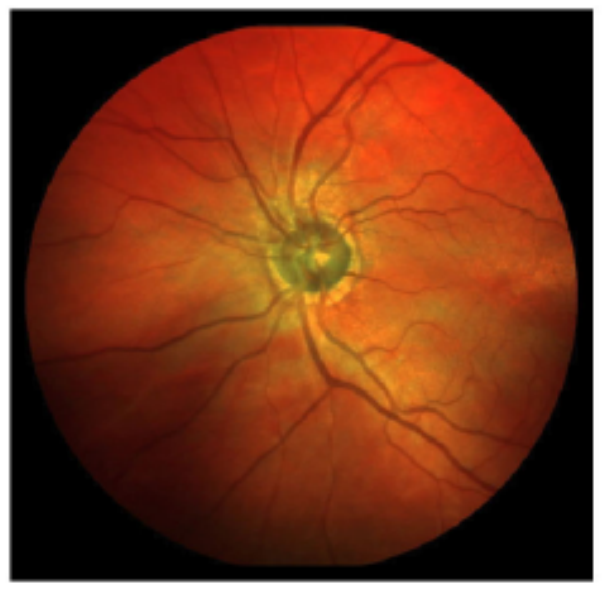

(a) Input ground truth.

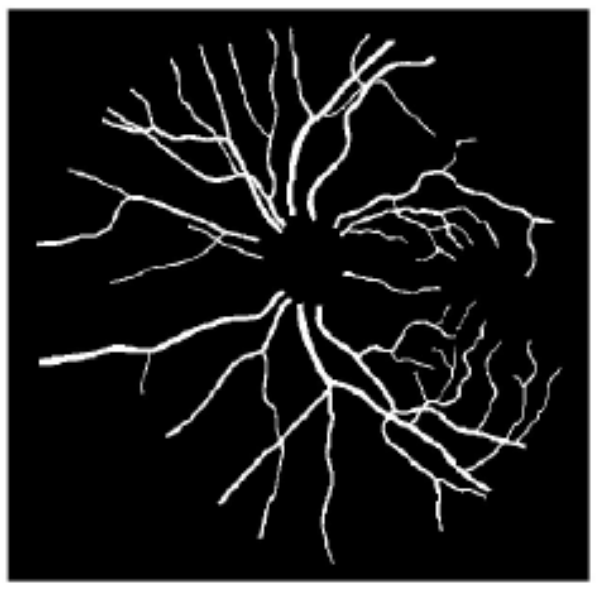

(b) Ground truth of (a).

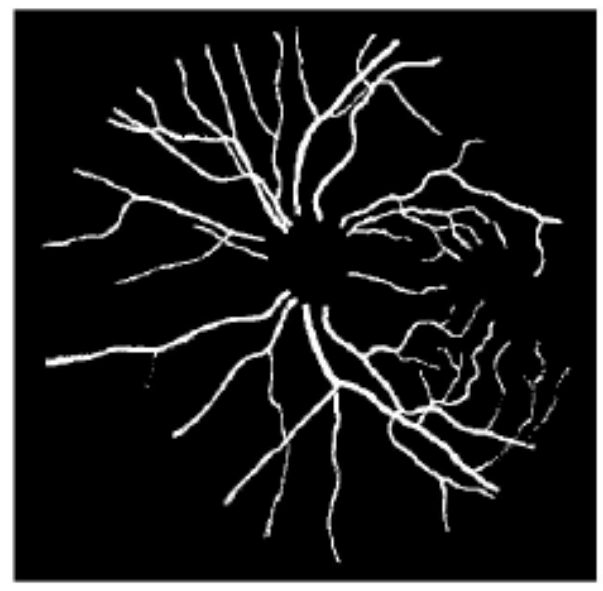

(c) Segmentation of (a) by ELEMENT.

Fig. 7. SLO images, (a) and (d), from the IOSTAR dataset. (b) and (e) show the ground truth of (a) and (d), respectively, while (c) and (f) are the results of ELEMENT's vessel segmentation, respectively.

of 45 degrees. The dataset contains a total of 24 images. The first image in the folder was used for training purposes, and segmentations were produced in the testing phase of the system for the remaining 23 images of the IOSTAR dataset. This dataset offers another type of retinal modality in order to evaluate the robustness of the proposed segmentation framework.

Table VII compares the performance statistics obtained by the ELEMENT framework with state-of-the-art methods on the IOSTAR dataset. Moreover, Fig. 7 shows some segmentation results produced with the proposed approach and the associated ground truth.

This experiment also highlights the robustness of the approach and how it can be easily adapted to different ocular modalities. In this case, the results were also better than the state-of-the-art in 3 out of 4 indices, as shown in Table VII.

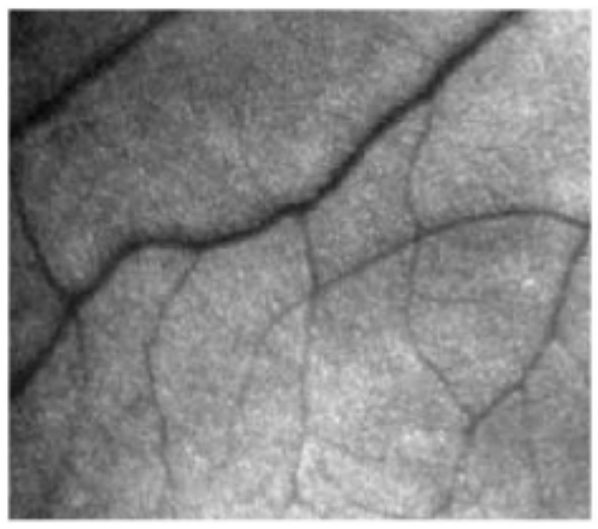

(a) Image 05-ODC-Patch1.

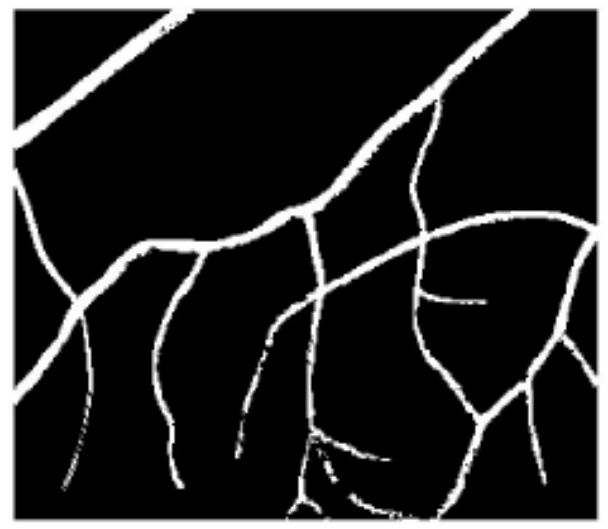

(b) Segmentation result of (a), obtained with ELEMENT.

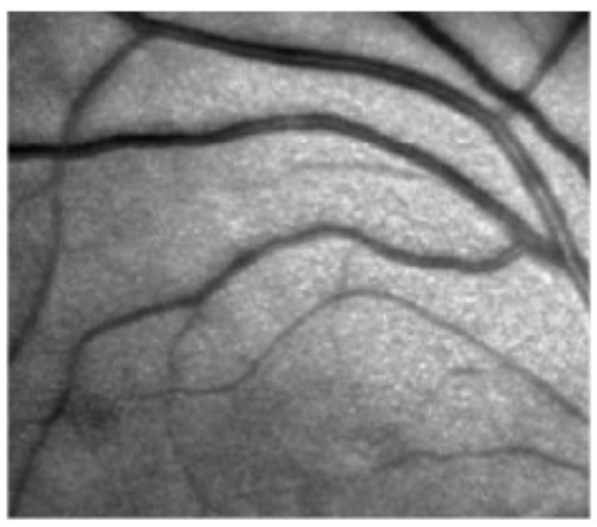

(c) Image 05-ODC-Patch2.

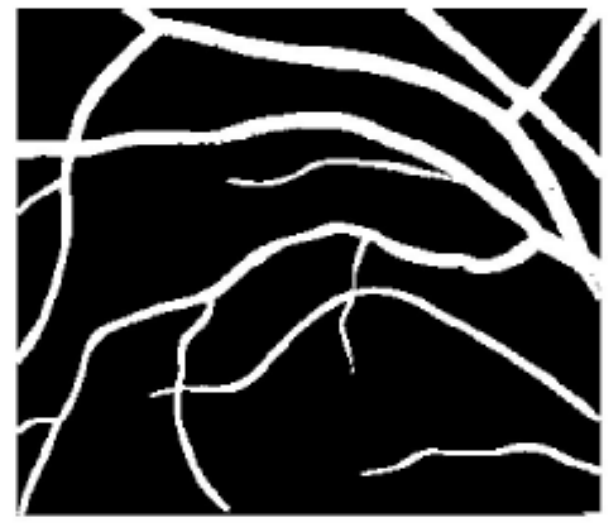

(d) Segmentation result of (c), obtained with ELEMENT.

Fig. 8. RC-SLO images and respective ELEMENT segmentation results.

### F. SLO Retinal Fundus (RC-SLO Dataset)

Finally, we provide a further last experiment with another SLO dataset, i.e., RC-SLO, provided by [46]. This dataset contains SLO image patches such as the ones shown in Fig. 8 along with the associated ELEMENT segmentation results.

The images in the RC-SLO dataset were acquired with an EasyScan camera (i-Optics Inc., the Netherlands), which is based on Scanning Laser Ophthalmoscopy. The RC-SLO dataset contains 40 image patches with a resolution of 360 x 320 pixels, annotated by experts. This dataset covers a wide range of challenging features, such as high curvature changes, central vessel reflex, micro-vessels, crossings/bifurcations and background artifacts. Table VIII shows a comparison of the ELEMENT results with the state-of-the-art methods.

### G. Feature Ranking and ROC Analysis

In this subsection, we provide two Receiver Receiver Operating Characteristic (ROC) curves for the best (IOSTAR) and worst cases (CHASE-DB) in the experimental studies. Fig. 9 shows in red the best case ROC, in blue the worst case ROC, and, in black the line of no discrimination (i.e., ZeroR). The ROC curve for the VAMPIRE dataset is very similar to that of IOSTAR. In summary, the ROC curves were better for the SLO and retinal fluorescein datasets in our experiments.

Finally, to shed light into the workings of the classifier and the individual importance of features, feature ranking is presented. Table IX shows feature rankings, which is associated to their relative importance in the classification task. The experiments were performed on all datasets and compared on 4 different

TABLE VIII
COMPARISON OF RESULTS WITH THE STATE-OF-THE-ART (RC-SLO DATASET)

| Work | TP | TN | ACC | AUC | F1 | MCC |
|---|---|---|---|---|---|---|
| Srinidhi et al., [51] | 84.88 | 96.66 | 95.81 | 0.9678 | - | 0.7029 |
| Zhang et al., [46] | 77.87 | 97.1 | 95.12 | 0.9626 | - | 0.7327 |
| Meyer et al. [80] | 80.9 | 97.94 | 96.23 | 0.9807 | - | 0.792 |
| ELEMENT | **95.88** | **98.62** | **98.35** | **0.9956** | **0.9187** | **0.9109** |

* TP refers to true positive rate and TN refers to true negative rate. * Accuracy (ACC) is given by: (true positives + true negatives) / total population. * AUC refers to the area under the Receiver Operating Characteristic (ROC) curve. * F1 refers to the F1-score. * MCC refers to the Matthews Correlation Coefficient.

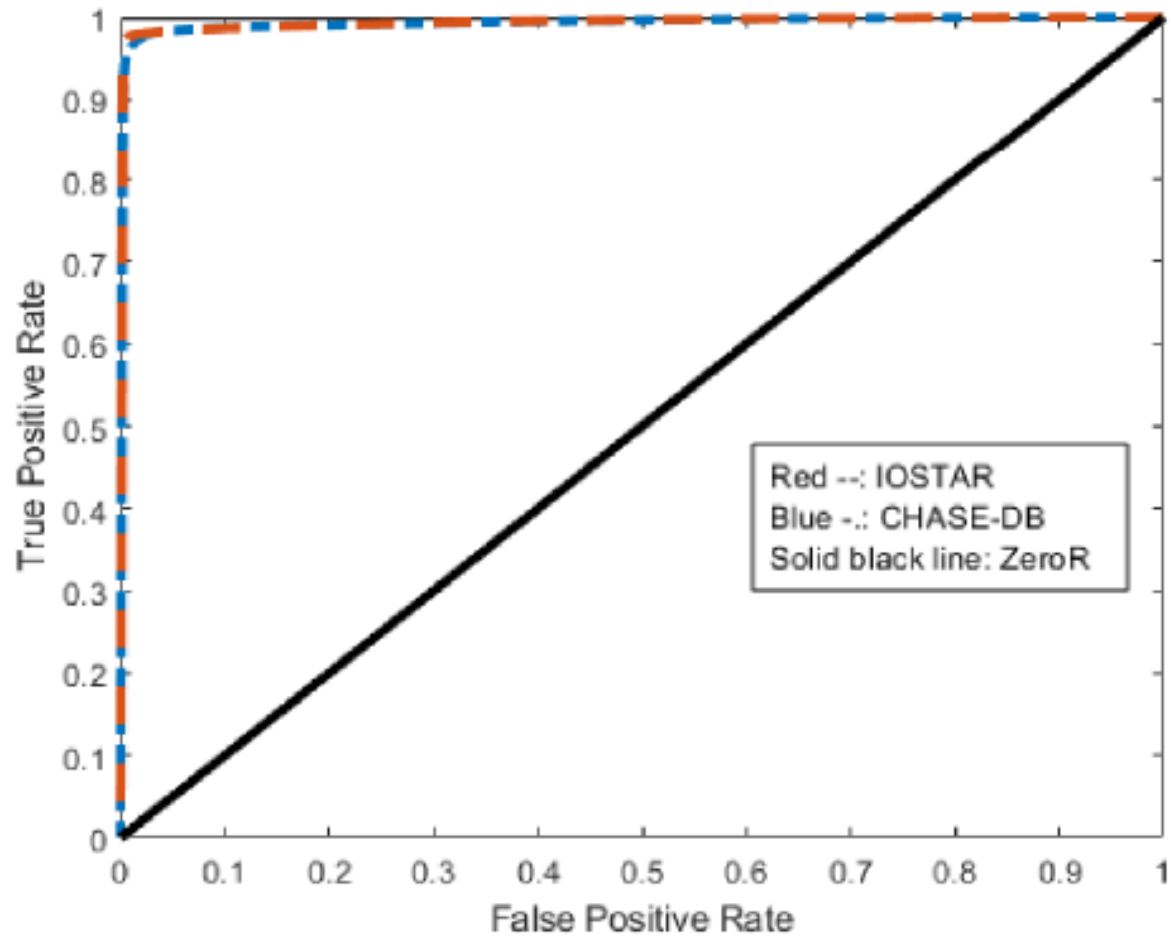

Fig. 9. ROC curves for the best and worst performances of ELEMENT on the retinal image datasets.

classifiers. This ranking was generated using Weka's Classifier Attribute Evaluator, which measures how the individual exclusion and inclusion of features impact on the classifier predictability performance.

We include the feature ranking position for the features that had a positive ranking impact according to the attribute evaluator (i.e., 1 is the first and best position). However, features that did not reach a positive ranking in some occasions (which are marked with a hyphen "-") may still be useful for some classifiers, and therefore we do not recommend removing these from the framework. In summary, all features showed some contribution to the classification task in general.

This experiment also reinforces how connectivity features play a very important role in the classification task, as shown in Table IX. Connectivity features were top 1 for every dataset and classifier. Connectivity encapsulates the information of the surrounding vessel pixels and assists the classifier in maintaining the tracing momentum of the vessels.

Features based on the Hessian matrix performed better than the Frangi filter itself, however this may be associated with the number of extracted features. For instance, in the case of the Frangi filter, we extracted a total of 3 features, while 10 features were extracted from the Hessian. Fine-tuning of the Frangi filter and a wider parameter selection may improve its ranking. Both

TABLE IX

FEATURE RANKING USING WEKA'S ATTRIBUTE EVALUATOR

| Features | DRIVE | | | | STARE | | | | CHASE-DB | | | | VAMPIRE | | | | IOSTAR | | | | RC-SLO | | | | Avg. Rank | Freq. |
|---|---|---|---|---|---|---|---|---|---|---|---|---|---|---|---|---|---|---|---|---|---|---|---|---|---|---|
| | J48 | RF | RBF | REP | J48 | RF | RBF | REP | J48 | RF | RBF | REP | J48 | RF | RBF | REP | J48 | RF | RBF | REP | J48 | RF | RBF | REP | - | - |
| Connectivity features (newly proposed) | 1 | 1 | 1 | 1 | 1 | 1 | 1 | 1 | 1 | 1 | 1 | 1 | 1 | 1 | 1 | 1 | 1 | 1 | 1 | 1 | 1 | 1 | 1 | 1 | **1** | 24 |
| Hessian | 2 | 2 | 2 | 2 | 2 | - | - | - | 2 | 2 | 2 | 3 | 2 | 2 | 2 | 2 | 3 | 3 | 2 | 3 | 7 | - | 7 | 7 | 2.95 | 20 |
| Sharpness | 3 | 4 | - | 3 | - | 2 | - | 2 | - | - | - | - | 3 | 3 | 3 | 3 | 4 | 4 | 3 | 4 | 8 | 7 | 8 | 8 | 4.23 | 17 |
| Frangi | 4 | 3 | - | 4 | 4 | - | - | 6 | - | - | - | 2 | - | - | - | - | 2 | 2 | - | 2 | 2 | 5 | 2 | 2 | 3.07 | 13 |
| Laplacian | 5 | 5 | - | 5 | 3 | 3 | - | 3 | - | - | - | - | 4 | 4 | 4 | 4 | 5 | | - | 5 | 9 | - | 9 | - | 4.85 | 14 |
| Gradients | 6 | - | - | 6 | - | 5 | - | 4 | - | - | - | - | 5 | - | 6 | 5 | - | - | - | - | - | - | - | - | 5.28 | 7 |
| Anisotropic diffusion | - | 6 | - | 7 | - | - | - | - | - | - | - | - | 6 | 5 | 7 | 6 | - | - | - | - | 4 | 3 | 4 | 4 | 5.2 | 10 |
| Morphological operations | - | - | - | - | - | 6 | - | - | - | - | - | - | 8 | 6 | 8 | 8 | - | - | - | - | 3 | 2 | 3 | 3 | 5.22 | 9 |
| Blur | - | - | - | - | - | 4 | - | 5 | - | - | - | - | 9 | 7 | 9 | 9 | - | - | - | - | 6 | 6 | 6 | 6 | 6.7 | 10 |
| Pixel value | - | - | - | - | - | - | - | - | - | - | - | - | 7 | - | 5 | 7 | - | - | - | - | 5 | 4 | 5 | 5 | 5.42 | 7 |

*J48 refers to J48Graft (similar to C4.5), RF refers to Random Forest, RBF refers to Radial Basis Function (RBF) Classifier and REP refers to Reduced Error Pruning (REP) Tree or simply REPTree.

*Freq. refers to the frequency that the feature appeared in the ranking logs as a positive contribution across the respective line.

*Avg. Rank refers to the average ranking across the respective line.

Frangi- and Hessian-based features appear to contribute more to the classification than the remaining features, as both reached an average rank close to 3, while the remaining non-connectivity features were at ranks greater than 4.

## V. DISCUSSION

Differences among the results obtained with each classifier, as a general rule, initially reflect their predictability potential. Classifiers are built upon distinct concepts, principles and theories and therefore, their internal behavior is specific to these definitions. Some classifiers work better with certain types of data than others. Algorithms such as REPTree and Hoeffding Tree are based on decision trees. Random Forest, which achieved promising results, is an ensemble classifier that uses several weak decision-tree learners. The promising results obtained with Random Forest may be associated to its ensemble process and this may be an unfair comparison. Still, we propose the use of this classifier in vessel segmentation applications, as it demonstrates the ability to better adapt to the specifics of the dataset.

We observed that all of the proposed features provide some degree of contribution in the classification process. Their individual contributions vary depending on the specifics of the dataset, as shown in Table IX. As a remark, we experimented with additional features, i.e., the Gabor and Kuwahara filters, gradients of higher order, the co-occurrence matrix, and features computed from the co-occurrence matrix generated on top of the Hessian matrix, however they produced negative or no impact on the classification performance, and were thus discarded from inclusion in this work.

It is noteworthy that ELEMENT outperformed 6 works that used deep learning methods on the DRIVE dataset [19]–[24]. We believe that this promising performance is due to the connectivity information that translates into coupling the advantages of machine learning-based classification and region growing. Conventional convolutional neural networks do not consider connectivity information by default, but rather they need to be tweaked to work with this kind of information (like the novel GNN networks).

As a final remark, a single text file containing the extracted features of a single image from the DRIVE dataset consumes approximately 160 megabytes. Experiments were ran using an Intel I7-7700HQ CPU clocked at 2.8GHz (fixed - no turbo boost) and 8 gigabytes of memory. The segmentation results, source code, ground truth and extracted features can be found at [75].

## VI. CONCLUSION

This work proposes ELEMENT, an adaptive vessel segmentation framework, which includes a novel set of features for vessel segmentation and a new means of classification that adheres to principles of region growing. The methodology was tested against three imaging modalities, namely, fundus photography, fluorescein angiography and scaning laser ophthalmoscope.

Connectivity features coupled with information that accounts for ridges and grey-level/texture information, and a specific pattern of growth for vessel pixel detection, provide a robust and informative feature vector for pixel classification, leading to accurate vessel segmentation. ELEMENT produced the top results in the published literature for the SLO and FA datasets and very competitive results in the case of the retinal fundus image datasets (i.e., best performance in the case of STARE and CHASE-DB datasets) in terms of accuracy, true positive rates and AUC.

Future avenues for research include the application of this framework for vessel segmentation in MRI datasets and the further extension to 3D vessel segmentation by suitable augmentation of the feature set. This is yet another scenario where connectivity features coupled with region growing could prove valuable in terms of classification performance.

3D vessel segmentation can be achieved by considering 3D connectivity features, which is a straightforward extension of the 2D connectivity features. Some grey-level features would also require adaptation to 3D space. In order to overcome this, a possible and simple approach is to include the z-axis (depth) index of the voxel as a feature. This could enable classifiers to separate and delineate the model according to the depth of the slice in the 3D space.

Furthermore, we also plan to develop automated algorithms for the detection of aneurysms and stenoses. Restriction in blood flow can be identified using heuristics once the vessels are segmented from the background structures, which is the output of this work. Sudden and unexpected reductions in the segmented vessel area may indicate blood flow restrictions. Along this line of thought, we plan to train machine learning algorithms to segment vessel structures (this work) and to later identify such restriction points, so as to quantify the percentage of blockages.